\documentclass[12pt,twoside]{skoltech_thesis}
\usepackage{subcaption}
\usepackage{amsmath}
\usepackage{amssymb}
\usepackage{graphicx}
\usepackage{algorithm}
\usepackage{algcompatible}
\usepackage{wrapfig}

\algnewcommand\INPUT{\item[\textbf{Input:}]}%
\algnewcommand\OUTPUT{\item[\textbf{Output:}]}%

\begin{document}

\title{CNN with large memory layers}

\author{Rasul Karimov}

\degree{Data Analysis and Management (Computer Science and Engineering)}

\degreemonth{June}
\degreeyear{2020}
\thesisdate{May, 2020}


\supervisor{Victor Lempitsky}{Associate Professor, PhD}
\cosupervisor{Yury Malkov}{PhD}

\chairman{TODO}{TODO, Skoltech}

\maketitle



\setcounter{savepage}{\thepage}
\begin{abstractpage}
This work is centred around the recently proposed product key memory structure \cite{large_memory}, implemented for a number of computer vision applications. The memory structure can be regarded as a simple computation primitive suitable to be augmented to nearly all neural network architectures. The memory block allows implementing sparse access to memory with square root complexity scaling with respect to the memory capacity. The latter scaling is possible due to the incorporation of Cartesian product space decomposition of the key space for the nearest neighbour search. We have tested the memory layer on the classification, image reconstruction and relocalization problems and found that for some of those, the memory layers can provide significant speed/accuracy improvement with the high utilization of the key-value elements, while others require more careful fine-tuning and suffer from dying keys. To tackle the later problem we have introduced a 
simple technique of memory re-initialization which helps us to eliminate unused key-value pairs from the memory and engage them in training again. We have conducted various experiments and got improvements in speed and accuracy for classification and PoseNet relocalization models.

We showed that the re-initialization has a huge impact on a toy example of randomly labeled data and observed some gains in performance on the image classification task. We have also demonstrated the generalization property perseverance of the large memory layers on the relocalization problem, while observing the spatial correlations between the images and the selected memory cells.
\end{abstractpage}


\section*{Acknowledgments}
We wish to express our deep sense of gratitude and profound thanks to Karim Iskakov and all the engineers in Samsung AI Moscow who contributed to the project in one way or another. We are hugely indebted to Samsung Research Center for the 
provided resources that gave us the chance to implement the models and conduct the required experiments.


\pagestyle{plain}
\tableofcontents
\newpage
\listoffigures
\newpage
\listoftables

\chapter{Introduction}
With the huge development of deep learning, neural networks have made significant progress in 
various tasks such as image classification \cite{imagenet2012}, speech recognition 
\cite{graves2013speech}, and machine translation \cite{seq2seq}. As it was shown in 
\cite{spigler2019jamming, neyshabur2018towards}, with sufficiently large data, 
increasing the capacity of neural networks could lead to far superior prediction accuracy. Therefore,  
scaling of both training and model size has been a central problem of deep learning field in recent 
years. For a typical neural model where the single input data sample depends on all the parameters 
of the network, therefore increasing both dataset and model sizes leads to a nearly quadratic 
surge in the training costs \cite{shazeer2017large}.

Previous works \cite{davis2013low, cho2014cap} proposed several ideas of increasing the model 
size with no proportional increase in a computational complexity.
Most of the proposed models rely on 
the sparse gating mechanism which determines whenever a particular node of the graph should be calculated during the forward pass of the network. 
This is the type of a branching problem with a discrete decision which is 
being solved with 
REINFORCE algorithm \cite{sutton2000reinforce}. It was applied in \cite{bengio2015conditional} and gave good results on \texttt{MNIST} dataset and \texttt{CIFAR-10} \cite{Krizhevsky09learningmultiple} with a reasonable speed up.

Instead of REINFORCE estimator, one can also apply the ideas from \cite{maddison2016concrete,
jang2017gumbel} by relaxing the discrete skipping decisions with reparametrization technique adoption. However, 
these approaches usually find sub-optimal solutions due to the approximation error introduced by the 
reparametrization trick \cite{wang2018skipnet, abati2020conditional}.

Other approaches rely on learning binary masks as the sparse \(l_{0}\) regularization term for the final 
objective. Works like \cite{nagel2020adaptive} employ a \textit{rectified sigmoid} proposed in \cite{louizos2018learning}
to learn binary decision choices. Authors apply regularization during post-processing to quantize the weights of the 
network, but the idea could be used in the training phase too. Recently, the paper \cite{popov2020neural} on differentiable
tabular data with neural networks has leveraged the \textit{entmax} transformation \cite{niculae2018sparsemap} to learn "hard" decisions for binary branching problem in decision trees.

Though solving the issue of scalability, models still fall short in giving promising results due to 
the following challenges:
\begin{itemize}
  \item GPUs are optimized to work faster with arithmetic tasks rather than branching.
  \item Batching reduces the batch sizes for conditionally activated chunks, therefore 
  complicating parallelization.
  \item REINFORCE estimator has a large variance, making it hard to get a strong bias during train. 
  There are some variance reduction tricks \cite{greensmith2002vreduction, mao2018variance} that try to solve the issue but most of them skew the bias-variance trade-off with no tuning on hyperparameters applied.
  \item Nearly all the methods suffer from neuron dying problem - if at some moment of the training a gate is not open for any input sample, this means it is highly unlikely it will be open at any further moment since the gate receives only zero gradient. 
\end{itemize}

A recent work \cite{large_memory} on the over-parametrized language models, on the other hand, rely on the key-value memory structures to scale the set of parameters in the neural network. Authors rely on the product key space decomposition for nearest neighbours to scale the networks with a little or no change in the performance and the memory consumption. These results encouraged us to research these methods in 
the computer vision applications.

As an extension of \cite{large_memory}, we augment the product key layer with the key-value re-initialization mechanism which allows to solve the dying neuron problem. The mechanism is based on re-initialization of dead or underutilized keys-values pairs using the information from more successful key-values pairs.

\chapter{Related work}
\textbf{Memory layers in neural models}
\textit{Memory augmented neural networks} (MANNs) augment neural networks with external memory 
block which allows for explicit access with differentiable write and read operations. The memory is 
usually implemented as a \textit{key-value differentiable dictionary} 
\cite{pritzel2017nec}, \textit{array-structured
memory} such as Turing machine (NTM) \cite{turing2014}, or recently proposed product-key memory layers 
\cite{large_memory}.

Key-value memory architectures were analyzed extensively in deep learning literature. Mostly the models 
are based on the architectural designs described in \cite{memory2015, report-memory-2015} and used mainly 
in natural language processing field (NLP) such as document reading and question answering tasks. The 
key-value paired structure can be seen as a generalization of how the huge context for question answering is stored 
in the memory. That makes key-value memory a natural choice for these tasks. And with the recent 
advancements in attention models (AM) \cite{bandanau2014am, attentionneed}, it is becoming the predominant concept 
in natural language processing literature with the usage case in nearly every NLP task. 

The key-value 
design structure is a sparse version of attention models, and as previously described in 
\cite{cordonnier2020On, bello2019attention} the self-attention mechanism could be competitive in replacing convlolutions as a computation primitive for object detection and image classification. We hope to leverage and improve upon those ideas for computer vision applications.

There were also some works in extending the key-value structure, \cite{cache2017} using unbound cache model 
to provide better modelling of infrequent words that occur in a recent context, including the 
out-of-vocabulary words. This is the evidence of interpretability of learned key-value structure which 
provides the linear interpolation between the values of the table. Other works \cite{khandelwal2020generalization} 
focus on the interpretability of memory blocks by linearly interpolating baseline language models (LM) with 
\textit{k}-nearest neighbours (\textit{k}-NN) models and assessing the generalization and memorization 
capability of the final models. Authors report increased perplexity in all of the experiments.

Other approaches have successfully applied memory layers to store image features, 
more specifically for image 
captioning \cite{park2017attend}, image generation tasks \cite{kim2018memorization}, and 
video summarization \cite{lee2018summarization}.

Some neural network designs include non-differentiable memory layers with unsupervised updates. 
These models mostly rely on the architectural ideas of \cite{lukasz2017memory} and rely on contrastive 
loss to update the memory block. Authors of \cite{lukasz2017memory} have demonstrated the 
the efficiency of their memory block in few-shot and one-shot learning tasks, while \cite{yoo2019color} 
has shown the advantage of using the memory in style transfer \cite{gatys2015style} tasks with limited data.
While in supervised approaches of memory usage where we are learning the mapping function between two spaces, 
in the unsupervised approach memory block is used for storing latent vectors with the ability to interpolate 
between them. This is the important property of memory blocks that is
implicitly used in most of the models.

Some works incorporate memory-like structure to store a bank of weights accessing them sparsely and using 
\textit{k}-nearest neighbours to retrieve a set of indices that allows to encode the feature vector 
into the discrete code. There are some promising results in auto-regressive models \cite{oord2017neural} 
giving high-fidelity image reconstruction results in \cite{oord2018neural}. Authors argue that the discrete representation is a more natural fit for complex reasoning, planning and predictive learning.

Moreover, memory layers were successfully incorporated in graph knowledge representation problems, with 
promising results in graph classification tasks \cite{graph_metric}.

\textbf{Learning compact representations, ANN}. Since the task of exact top-k neighbour search is expensive 
in practice for high dimensional embeddings, i.e. linear scale to the search set size, 
practitioners in the field usually resort to more scalable \textit{approximate nearest neighbours} methods  
(ANN). Popular methods include
\textit{Locality sensitive hashing} (LSH) \cite{lsh-1999, lsh-nips} that relies on random space partitioning, graph methods like \textit{Navigable small world graphs} (NSW) \cite{malkov2014} and Hierarchical NSW 
(HNSW) \cite{baranchuk2018} based on constructing the search algorithm by finding the clusters of data.
Another important subset of methods utilize \textit{quantization} \cite{ge2013, jegou2011, babenko2014inverted} to reduce the memory consumption and speed up the distance calculations. Many of those methods exploit the idea of product decomposition, e.g. assumption that the space can be decomposed into a Cartesian product of low dimensional spaces. 

\textbf{Product decomposition in neural models} 
Most of this thesis is inspired by the work of \cite{large_memory} which are showing the efficiency of using 
product key memory layer in language modelling tasks. Here product key is a structure that allows more efficient scaling by scarifying some expressiveness of the model. Authors find that the language model 
augmented with memory with only 12 layers can outperform in accuracy a baseline transformer model with 24 
layers while giving two times less inference time.
The \cite{large_memory} didn't however address the problem of dying keys other than by adding noise to the query (via batchnorm) and was focused solely on NLP applications. 

Product quantization in general has also been used in many computer vision applications, starting from 
scalable supervised \cite{ning2016scalable} to semi-supervised \cite{jang2020generalized} image retrieval 
tasks. There are some promising results \cite{yi2020face} in face video generation with memory augmented 
Generative adversarial networks (GAN) \cite{gans2014}. 

\textbf{Classification networks}
Huge chunk of work \cite{googlenet, he2016deep, krizhevsky2012deep, simonyan2014very} is done in designing the neural networks for the image classification problems. In our experiments we mainly focus on the \texttt{ResNet}-like \cite{he2016deep} networks. Some recent work \cite{tan2019efficient} demonstrated the 
SOTA results in \texttt{ImageNet-2012} dataset \cite{imagenet2012} with the help of the reinforcement learning to tune the models and the whole architecture. Most of the existing neural networks for image classification rely on the convolutional block but there were some recent works suggesting the self-attention mechanism with promising results \cite{bello2019attention, ramachandran2019stand}.
\chapter{Methods}
\section{Product key memory}
\subsection{Design overview}
The overall pipeline of the differentiable product memory layer is similar to most of the 
key-value data structures that are augmented into neural network models \cite{turing2014, report-memory-2015, memory2015}. 
More specifically, product memory design in our work is heavily inspired by previously 
proposed architecture in \cite{large_memory}. Here we build models upon this design to solve classification, regression, and reconstruction computer vision tasks.

\begin{figure}[h]
    \centering
    \includegraphics[width=\linewidth]{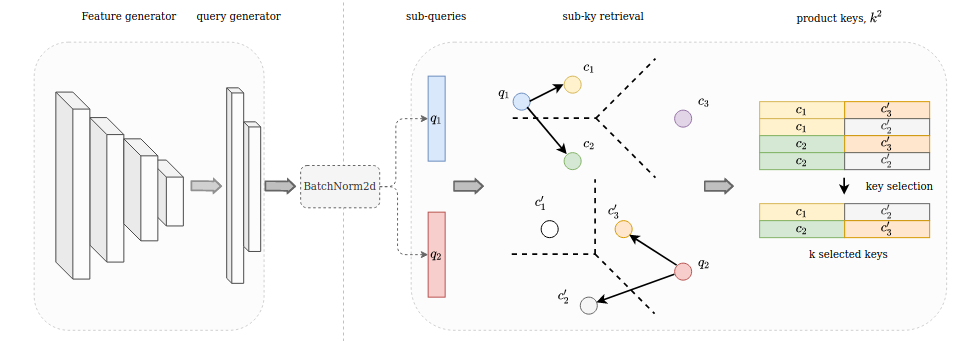}
    \caption{Overview of the product key-value memory design. Feature generator is the baseline
    image model which outputs the latent vector that is projected by the the query generator, normalized and
    divided into the two sub queries \(q_1\)/\(q_2\) to work with product key values. As the result we have \textit{k}
    selected keys and the corresponding values from \(\mathcal{V}\).}
    \label{fig:prod-key-memory}
\end{figure}

Higher view of the architecture is illustrated in Figure \ref{fig:prod-key-memory}. The central idea is to augment baseline 
convolutional neural networks with sparse memory access. The input of the memory layer is the latent feature vector 
that describes the given input image. Depending on where we place the memory layer, the query can represent features like 
brightness gradients or colours with more complex patterns in later layers \cite{zeiler2014vis}. Therefore, the choice of 
memory access placement is important. 
Given the input query, memory block finds the distance scores by comparing it with all of the keys in 
the table and selecting the values associated with top-\textit{k} distance scores. The scores are then used to produce the output 
\(m(x)\) via a weighted sum over the values associated with the selected keys:
\begin{equation*}
    m(x)=\sum_{i \in \mathcal{I}} w_{i} v_{i}
\end{equation*}
where \(\mathcal{I}\) is the set of top-\textit{k} indices by distance scores, \(w_{i}\) are the scores, and \(v_{i}\) are the values 
in the memory layer.

\textbf{Query generation.} The memory block consists of the query generation network which is a  learnable projection function \(q : x \rightarrow q(x) \in \mathbb{R}^{d_q}\) mapping the \textit{d}-dimensional input vector 
\textit{x} into the \textit{\(d_q\)}-dimensional query vector. 
Typical dimension sizes of the query vectors in our experiments are from 256 up to 2048.

Also, since the keys are initialized in a fixed range, we follow \cite{large_memory} adding \texttt{BatchNorm} \cite{batchnorm} 
layer on the top of the query network. This allows a better overlap between the distribution of keys and queries. And as in \cite{large_memory} we observe higher utilization of the memory with \texttt{BatchNorm} enabled.

\textbf{Key assignment} We have a resulting query \(q(x)\) 
that should be assigned to the closest keys with the selected metric for the distance. Let \(\mathcal{K}=\{k_{1}, \ldots, k
_{|\mathcal{K}|}\}\) is the set of all keys, the set is 
composed of \(|\mathcal{K}|\) \textit{\(d_q\)}-dimensional vectors that are uniformly initialized 
in the \(\mathbb{R}^{d_q}\) space. We can define the differentiable procedure to find a 
weighted sum over the value vectors (the memories associated with top-\textit{k} keys). The sum is weighted by the 
distance scores between the subset of the top-\textit{k} keys and the query value \(q(x)\). Top-\textit{k} procedure finds the most closest keys to the given query, i.e. maximization of the 
chosen similarity measure \(d(\cdot, \cdot)\). The overall algorithm is:
\begin{align*}
    & \mathcal{I} = \mathcal{T}_{k}\left(d(q(x), k_{i})\right)\\
    & w = \operatorname{softmax}\left(\left\{ d(q(x), k_{i})\right\}_{i \in \mathcal{I}}\right)\\
    & m(x) = \sum_{i \in \mathcal{I}} w_{i} v_{i}
\end{align*}
where \(\mathcal{T}_{k}\) denotes the top-\textit{k} operation which finds k largest values based on the 
similarity measure \(d(\cdot,\cdot)\). \(\mathcal{I}\) denotes the index set of the most similar keys to the 
query \(q(x)\) and \(w\) represents the normalized scores associated with the selected indices. The 
resulting value \(m(x)\) is the sum of the selected values weighted by the normalized scores. As we 
see due to the summation over the normalized values in the \(\operatorname{softmax}\) operation, 
the gradients can be calculated. Note that it is not possible to find the gradient for top-1 function, since there is no summation.

\textbf{Product keys} We see that the bottleneck of the given procedure is the calculation of the \(\mathcal{T}_{k}\) 
the operation which has linear complexity over the size of the key set \(\mathcal{K}\), so it
is hard to scale this procedure for large memory sizes. The remainder of 
operations are done for the reduced set of selected indices, e.g. the summation over 
top-\textit{k} normalized weight values. 

To solve the performance issue, authors of \cite{large_memory} propose to represents the key set 
in the form of two independent sets of half dimension size \(d_q/2\) vector sets \(\mathcal{K}_1\) 
and \(\mathcal{K}_2\) which constructs the Cartesian product set of resulting values with size \(|\mathcal{K}| = |\mathcal{K}_1| \times |\mathcal{K}_2|\). 
The query vector should also splitted into two sub-queries \(q_1\) and \(q_2\) to work in each of the key sets. 
We then find the closest keys in both sets as:
\begin{equation*}
    \mathcal{I}_{\mathcal{K}_{1}}=\mathcal{T}_{k}\left(\left\{ d(q_{1}(x), k_{i}^{1})\right\}_{i 
    \in\{1 \ldots|\mathcal{K}_{1}|\}}\right), \quad \mathcal{I}_{\mathcal{K}_{2}}=\mathcal{T}_{k}\left(\left\{d(q_{2}(x), k_{j}^{(2)})\right\}_{j \in\left\{1 \ldots\left|\mathcal{K}_2\right|\right\}}\right)
\end{equation*}

Then the two subsets of the keys associated with the index sets
\(\mathcal{I}_{\mathcal{K}_1}\) and \(\mathcal{I}_{\mathcal{K}_2}\) are multiplied together to form a new Cartesian product set. We are applying the top-\textit{k} operation on the newly created set and find the final subset of the top-\textit{k} keys.

\textbf{Choice of distance}
Authors in \cite{large_memory} experiment with the inner product as the single similarity measure 
for the provided experiments. We observe that using cosine similarity not only provides us with 
better numbers in some experiments but also gives us control over the selection process of 
the keys. Since the dot product is proportional to the vector norm, the key vectors with the largest 
vector lengths will be selected in most of the cases, while low norm vectors may be completely ignored. This means that the distance measure captures the most 
popular candidates, the latter can skew the similarity metric. We balance the skew by introducing the 
hyperparameter \(\alpha\) and raising the length to an exponent \(\alpha < 1\) to calculate the 
distance as:
\begin{equation}
    d_{cos}(q, k, \alpha) = |q|^{\alpha}|k|^{\alpha} \cos (\theta) = |q|^{\alpha}|k|^{\alpha} 
    \frac{q^{T} k}{|q| \cdot|k|}
\end{equation}

\textbf{Multi-head mode}
To make the model more expressive we are using the multi-head mechanism
\cite{attentionneed} which splits queries and keys into multiple chunks of vectors to be calculated independently. 
The similar calculations are conducted on each head and the results are concatenated. Due to the reduced dimension of each 
head, the overall computational complexity of the layer is similar to the single-head attention.

\textbf{Complexity}
Naive top-K key selection requires \(\mathcal{K}\) comparisons of \(d_q\) sized vectors, 
which gives us \(\mathcal{O}\left(|\mathcal{K}| \times d_{\mathrm{q}}\right)\) operations overall. 
When using the  product space \(\mathcal{K} = K_1 \times K_2\),
we have two separate sets of keys for subspaces with significantly reduced carnality  \(|K_1| = 
|K_2| = \sqrt{|\mathcal{K}|}\). The overall complexity for the first step then is: \(\mathcal{O}\left(
|K_1| \times d_q/2 + |K_2| \times d_q/2\right) = \mathcal{O}\left(|K_1| d_q\right) = \mathcal{O}
\left(\sqrt{\mathcal{|K|}} d_q\right)\). The second step is performed on the reduced subset of
\(k \times k\) elements so it will require \(\mathcal{O}\left(k^{2} \times d_1\right)\) operations.
Therefore overall complexity is:
\begin{equation*}
     \mathcal{O}\left(\left(\sqrt{|\mathcal{K}|}+k^{2}\right) \times d_{\mathrm{q}}\right)  
\end{equation*}
\section{Re-initialization trick}
\subsection{Overview}
While conducting our initial experiments on random data, we have observed that a toy neural network 
augmented with memory block struggles to fit the data with no multi-head mode enabled even though the model should have had enough capacity to fit the whole dataset. By conducting some ablation study and literature review
\cite{baan2019understanding} we have concluded that the problem is due to the correct 
initialization of the memory layer. Additionally, authors in \cite{voita2019multihead} suggest that 
most of the heads in the attention mechanism can be pruned without serious effect on the performance.
To tackle the initialization issues we are introducing the re-initialization trick that dynamically initializes 
unused keys during the training phase. We are describing the whole procedure below.

\begin{figure}[H]
    \centering
    \includegraphics[width=\linewidth]{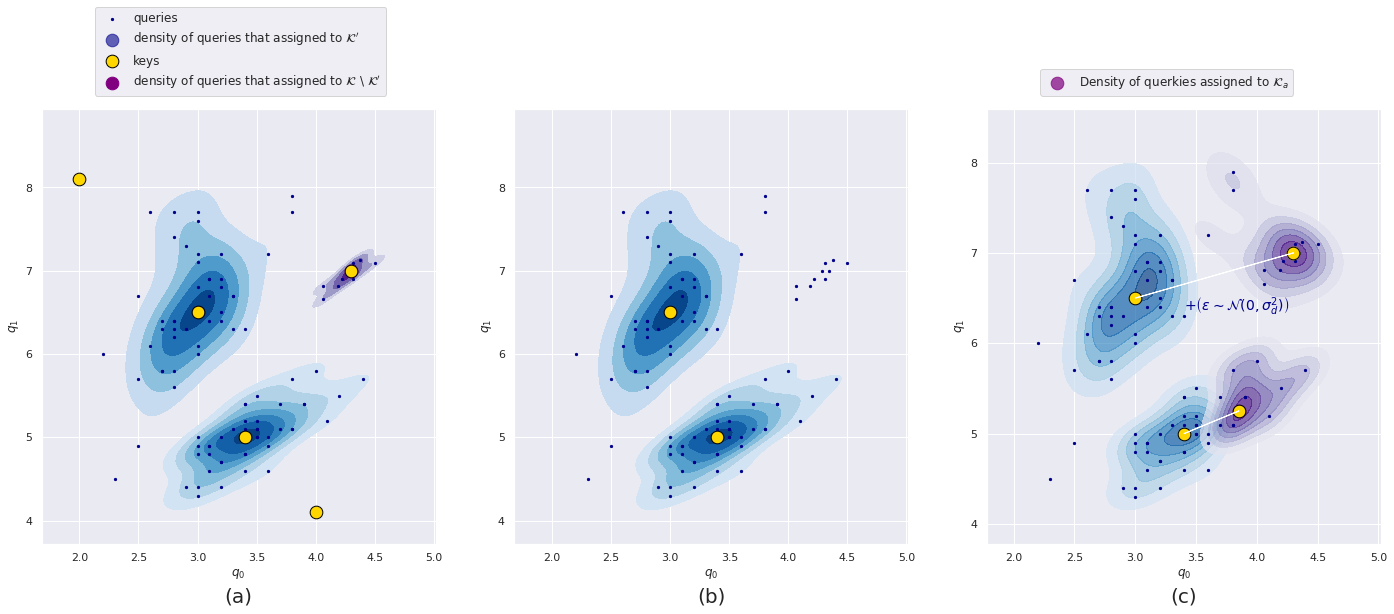}
    \caption{The illustration of the key re-initalization procedure. (a) We have 5 keys with three of
    which does not pass the threshold of being identified as the "utilized keys". We are removing those keys in the step (b). 
    And in the step (c) we are initializing new keys sampled from existing ones and perturbed with an
    additive Gaussian noise (\(\epsilon \sim \mathcal{N}(0, \sigma_d^2)\)).}
    \label{fig:experiment.explained}
\end{figure}

\subsection{Problem of dying keys}
Let's assume that we are working with the dataset of size \(|\mathcal{D}|\) which is equal to
the number of values in the memory  \(|\mathcal{M}|\). We could assume that augmenting the neural 
network with the memory layer could lead to the full convergence, i.e. perfect accuracy, because of 
the one-to-one mapping between the input and the memory elements. 
We hovewer did not observe this in our 
experiments with random data (description is in the experiments section), and classification tasks.
Instead, we discover continuously reduced cardinality of the selected key 
set at each iteration of the optimization, reaching some fixed value \(|\mathcal{K}^{\prime}|\):
\begin{align}
    & |\mathcal{K}^{\prime}| = \alpha |\mathcal{K}|, \, \alpha \ll 1 \label{eq:cardinality}\\
    & \mathcal{K}^{\prime} = \left\{ k_i \in \mathcal{K}| c_i > 0, c_i \in \mathcal{C} \right\}
\end{align}
where \(K^{\prime}\) is the set of selected keys during the inference, \(\mathcal{K}\) is the set of all keys,
and \(c_i\) is the utilization 
of the key \(k_i \in \mathcal{K}\) summed for the whole dataset, i.e. number of times 
the key \(k_i\) was selected. 
In the experiments we are not able to get full 
utilization of the selected keys and observed low final accuracy. We call this 
a problem of \textbf{dying keys}, when the optimizer is unable to pass gradients through certain 
key-value pair in the memory layer, leading us to the dead keys, useless for inference but still having a computational burden.

\subsection{Key re-sampling}
To solve the problem we implement a simple trick of key re-initialization, which is being executed 
during the training phase at certain points. We observe that during training, the key utilization converges to some specific number, as it is given in Equation (\ref{eq:cardinality}). 
We assume that the main reason for this is dying keys problem discussed in the previous section.
For this reason, we are running the pipeline 
of key re-initialization when the utilization plateau is reached. 

Here we describe the algorithm for a single product 
space key subset but the algorithm is applicable for both of key subsets. Let \(\mathcal{K}\) define 
the set of all the keys in memory and \(\mathcal{K^{\prime}}\) is the subset of utilized keys where 
\(|\mathcal{K^{\prime}}| \ll |\mathcal{K}|\). We also introduce the hyperparameter \(k_a\) which 
will control how many keys should be re-sampled at the each call of our key initialization procedure. 
Then we have:
\begin{align*}
    & \mathcal{I}_a =  \left\{ i|i \sim U\{0, |\mathcal{K}^{\prime}|\}\right\}\\
    & \mathcal{K}_{a} = \left\{ k_i \in \mathcal{K}^{\prime}| i \in \mathcal{I}_a  \right\} + 
    \epsilon, \epsilon \sim \mathcal{N}(0, \sigma_{n}^{2})\\
    & \mathcal{K} = \mathcal{K}_{a} \cup \mathcal{K}^{\prime}\\
\end{align*}
where \(\mathcal{I}_a\) is the set of indices sampled uniformly from the used keys \(\mathcal{K}^{\prime}\), 
\(\mathcal{K}_a\) is the sampled set of utilized keys perturbed with guassian additive noise. We have an additional 
hyper parameter \(\sigma_n\) that controls the noise variance of the re-initialized keys, i.e. the magnitude of 
difference between the original keys and re-initialized ones.
Then the existing set of utilized 
keys are expanded by \(\mathcal{K}_a\). 
The sampling mechanism we discussed above is very basic, but sampling more from the regions of high density/low density could potentially bring us more gain both in prediction accuracy and the compactness
of the final representation. This, however, requires the re-initialization algorithm to be able to sample key points in 
the regions with higher density. Something like rejection sampling algorithms, i.e. \textit{Metroplolis-Hastings} 
algorithm \cite{metropolis} could save us here, by defining the multimodal normal distribution and the utilization of the key values as the mean parameter. But because of the difficulty of tuning the rejection 
sampling algorithm, we plan to test those algorithm in the future and resorting to simple re-sorting discussed in the following section.

\subsection{Re-sorting and key-value reinititalization}
To give more priority on the regions of high density during sampling, we are sorting the keys in the set by 
the utilization coefficients \(c_i \in \mathcal{C}\), and adapting a naive thresholding logic to 
eliminate the least utilized keys by removing those with the values less than the hyper parameter \(d_k\)
Then the index set calculated is:
\begin{align*}
    & \mathcal{K}_{d_k} = \left\{ k_i \geq d_k | k_i \in \mathcal{K} \right\} \\
    & \mathcal{I}_a =  \left\{ i \in \{0..|K|\}| k_i \in \mathcal{K}_{d_k} \right\}
\end{align*}
After we resample the keys by eliminating the least utilized, we to initialize new values 
\(\mathcal{U}_{a}\) that will be mapped to the elements 
of the Cartesian product set of the new keys \(\mathcal{K}_a\). Because of the set product, adding single key 
to the subset will add \(|\mathcal{K}_a| |\mathcal{K}|\) 
new values into the memory. For each key from the first product set, we are initializing new values associated with 
the resulting keys concatenated with the given key from the first set and all the existing keys in the second set.
The same applies to the values associated with the second product set.  The overall algorithm for 
the re-sampling step is demonstrated in Algo \ref{algo:solution}.

\begin{algorithm}
    \caption{Re-initialization algorithm}
  \begin{algorithmic}[1]
    \INPUT existing set of keys \(\mathcal{K}^{\{0,1\}}\), noise variance \(\sigma_n\), utilization set
    \(\mathcal{C}^{\{0,1\}}\), threshold parameter \(d_k\), and the weight values \(\mathcal{U}\)
    \OUTPUT Reinitialized set of keys \(\mathcal{K}\), Reinitialized values \(\mathcal{U}\)
    \STATE \textbf{function} keysort(\(\mathcal{K}, d_k\))\\
        \quad \(\mathcal{K}_{d_k} = \left\{ k_i \geq d_k | k_i \in \mathcal{K} \right\}\)\\
        \quad \textbf{return} \(\mathcal{K}_{d_k}\)
    \STATE \textbf{end function}
    \FOR{\(j \in \{0, 1\}\)} 
    \STATE \( \mathcal{K}_{d_k}^{j} = \operatorname{keysort}(\mathcal{K}^{j}, d_k)\)
    \STATE \( \mathcal{U} = \mathcal{U} \setminus \left\{ \mathcal{U}_i | i \in \left(\mathcal{K}^{j} 
    \setminus K_{d_k}^{j}\right) \times \mathcal{K}^{(-j)} \right\} \)
    \STATE \( \mathcal{I}_a^j \sim U\left\{ i \in \{0..|\mathcal{K}^{j}|\}| k_i \geq 
    \mathcal{K}^{j}_{d_k} \right\} \) \, \textit{sample \(a\) indices from the discrete distribution}
    \STATE \( \mathcal{K}_{a}^{j} = \left\{ k_i \in \mathcal{K}^{j}| i \in \mathcal{I}_a^{j}  \right\} + 
    \epsilon, \epsilon \sim \mathcal{N}(0, \sigma_{n}^{2}) \)
    \STATE \(\mathcal{K}^{j} = \mathcal{K}^{j}_{a} \cup \mathcal{K}^{j}_{d_k}\)
    \STATE \( \mathcal{U}^{\prime} = \left\{ u_i^{\prime} \sim U | i \in 
    [0..|\mathcal{K}_a^{j}| |\mathcal{K}^{-j}|] \right\}\)
    \STATE \(\mathcal{U}^j = \mathcal{U}^{j} \cup \mathcal{U}_a^{j}\) \quad \textit{We need to
    associate the indices of newly created values with keys}
    \ENDFOR
    \STATE \( \textbf{return} \, \mathcal{K}^{\{0,1\}}, \mathcal{U} \)
  \end{algorithmic}
  \label{algo:solution}
\end{algorithm}

\subsection{Re-initialization complexity}
Taking the constant complexity of random number generation we can assume that the index sampling 
from discrete distribution is also constant. Then the complexity of key generation for both subsets 
is \(\mathcal{O}(d_q \times |\mathcal{K}_a|)\). The complexity for value re-sampling is \(\mathcal{O}(|\mathcal{K}_a| \times |\mathcal{K}| \times d_1)\) which in result give us the 
complexity of the whole procedure as: 
\begin{equation*}
    \mathcal{O}(d_q \times |\mathcal{K}_a|) + \mathcal{O}(|\mathcal{K}_a| \times |\mathcal{K}| \times d_{v})
\end{equation*}
where \(d_{v}\) is the dimension of the memory value vectors.

\section{Classification pipeline}

\begin{wrapfigure}{L}{0.3\textwidth}
\centering
\includegraphics[width=0.25\textwidth]{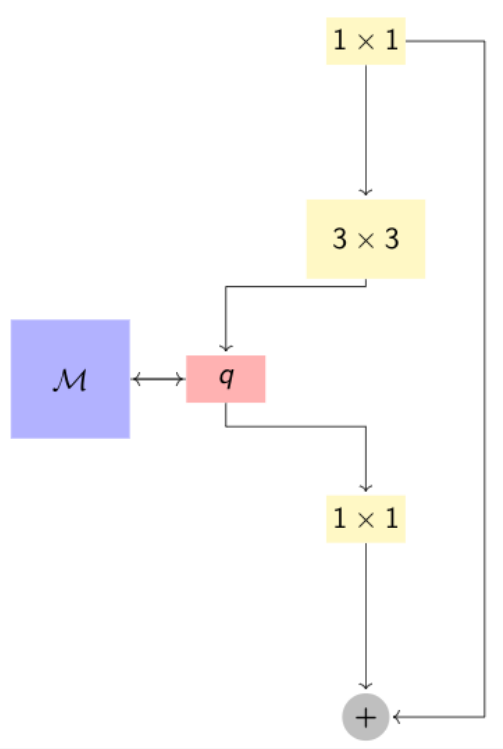}
\caption{\label{fig:bottle}Bottleneck \cite{he2016deep} augmented with a memory block \(\mathcal{M}\).}
\end{wrapfigure}
We are augmenting various types of the classification neural networks with the 
memory layer defined in the sections above. \texttt{ResNet} \cite{he2016deep} is the 
baseline architecture for most of the experiments.
The first idea is to augment the \texttt{Bottleneck} block with the memory layer.
The memory is inserted after the (\(3\times 3\)) kernel size 
convolution layer. We could also add the memory access before the 
middle convolution layer but we didn't find any differences between the two methods so we just stick with the first design. We are keeping the baseline high-level architecture 
the same by only replacing the single \texttt{Bottleneck} block with the 
augmented version. Replacing a single layer should be enough to observe the effect of the memory, while having only a single layer with relatively low spatial size allows less carrying about the efficiency of the layer implementation. 

Inspired by \cite{squeeze-e} we are also adding the memory access in squeeze-and-excitation (SE) like manner. SE is a computation block that can be built upon any transformation in the network. It  consists of two parts. First, \textbf{Squeeze} which, given the feature map, captures the global spatial information and \textit{squeeze} it into the channel descriptor. Authors use global average pooling with that goal. Second, \textbf{Excitation}, which employs the simple gating mechanism upon the projection of squeezed vector with a sigmoid activation. The projection is the bottleneck with two fully connected layers around the non-linearity.
The bottleneck layer successfully limits the complexity of the SE block by introducing the dimensionality-reduction with ratio \(r = d_{in} / d_{out}\), where \(d_{in}\) is the dimension size of input vector and \(d_{out}\) is the dimension size of the vector after the first projection in the bottleneck.

\begin{figure}
    \centering
    \includegraphics[width=\linewidth]{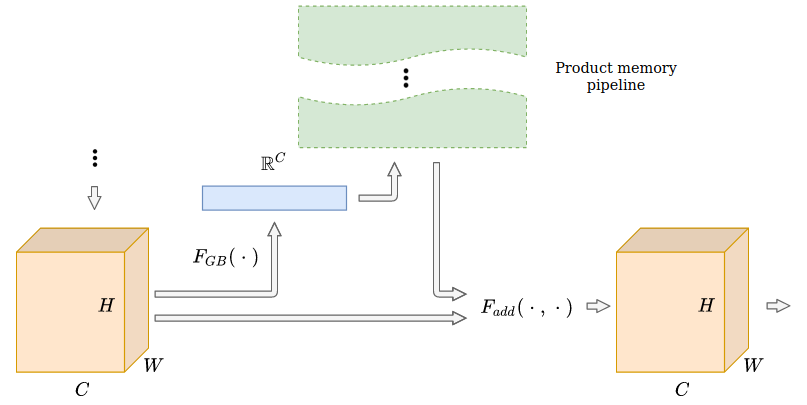}
    \caption{The overview of the modified squeeze-and-excitation block augmented with the memory layer.
    \(F_{GB}\) is the function of global pooling which reduces the dimension of 3 dimensional feature map to the signle dimension, and \(F_{add}\) is channel-wise addition of a resulting vector from memory to the original feature tensor.}
    \label{fig:squeeze_excitation}
\end{figure}

We setup nearly the same design but with three main differences, first, we are replacing only one block instead of every/several blocks in \cite{squeeze-e} (fewer SE blocks give worse final score). This 
reduces the number of parameters to be stored in the memory and the overall FLOPS required in the inference. Second, channel-wise 
feature response is fed to the memory instead of the MLP with two fully-connected (FC) layers around 
the non-linearity. This design helps us to tackle the issues 
of large spatial shapes of the query input and therefore softens the overall performance drop.
 Finally, instead  of re-scaling the values of the feature map with the gating output, we are simply adding the embedding 
pixel-wise, i.e. replacing multiplication by addition operation and adding the embedding to each pixel of the feature map. The overall model of memory augmented 
squeeze-and-excitation block is illustrated in Figure \ref{fig:squeeze_excitation}.

Another option is to simply add the memory block as an additional layer between the existing ones. This way we still have the issues with large spatial shapes, especially for the earlier layers. We are testing this 
design type with the \texttt{ImageNet} dataset \cite{imagenet2012}.

\section{Regression pipeline}
To test the capability of the memory layer to work on regression problems, we are also experimenting with the camera relocalization problem \cite{posenet}. The task is to find the camera extrinsics, i.e. the camera's location in the real world and its direction, from the single image. Inferring the camera extrinsics is a crucial task for mobile robotics, navigation, augmented reality.

Authors of the PoseNet neural network \cite{posenet} construct the simple neural model which 
consists of the backbone network and two additional layers that 
map the feature representation of the input into the pose and the direction values.
First, it is the regression feature generation network as a basis for the architecture. For 
that purpose \texttt{GoogleNet} \cite{googlenet} is used, we are replacing it with \texttt{ResNet}
\cite{he2016deep} to conduct our experiments. The output of the backbone is fed to the fully connected 
layer, which is then regressing values for direction and orientation separately. Authors of the paper suggest to parametrize the direction
with quaternions because of the overall simplicity compared to the rotational matrice, i.e. advantage in size: 4 scalars vs 9 in rotation matrix and speed since quaternion multiplication is much faster compared to a matrix-vector product used for rotation matrices. Additionally, since the rotation matrices \(n\times n\) are the members of \(SO(n)\) \cite{kendall2017geometric}, they have the orthonormality property that should be preserved during the optimization, which is generally the hard problem.

Since quaternion, 
\textbf{q}, is identical to \textbf{-q}, this leads us to the issue of non-injectivity of the 
rotation value. To solve it authors normalize a quaternion rotation to a unit length:
\begin{equation*}
    \mathcal{L}_{q}(I)=\left\|\mathbf{q}-\frac{\hat{\mathbf{q}}}{\|\hat{\mathbf{q}}\|}
    \right\|
\end{equation*}

For position loss, \(L_2\) Euclidean norm is used. Introducing scaling factor \(\beta\), we 
can balance the overall loss, by keeping expected values of two losses approximately equal. We 
are not trying to tune the scaling factor in our experiments since it is not the main direction 
of this research, but we still experiment with a large grid of hyperparameters including various values for the scaling factor. The overall objective loss function is:
\begin{equation*}
    \operatorname{loss}(I)=\|\hat{\mathbf{x}}-\mathbf{x}\|_{2}+\beta\left\|\hat{
    \mathbf{q}}-\frac{\mathbf{q}}{\|\mathbf{q}\|}\right\|_{2}
\end{equation*}
We are experimenting with memory block by replacing the fully connected layer before the final 
regressor of feature size 2048. Since the data size (King's College \cite{posenet}) on which the 
experiments are conducted is relatively small, we are constraining ourselves with setting the memory 
size to 1k/10k values. We also regularize the memory layer by augmenting weights with \texttt{Dropout} (multiplicative 
binomial noise) but find far worse results.

\section{Image reconstruction pipeline}

To test the memory layers further we are working with an image reconstruction problem on the \texttt{Imagenet-2012} \cite{imagenet2012} dataset. Image reconstruction is the type of 
dimensionality reduction problem to learn the 
projection function that could inject the given image into the 
latent representation in the compact manifold (data encoding) and then generate the image from the given latent. Autoencoder is a neural approach that helps us to tackle the problem in an unsupervised fashion. In the basic design of the autoencoders, we have two networks: encoder which maps the image into the small latent code and a decoder which generates the image from the code.

We are experimenting with several autoencoder designs but stick to:
\texttt{DCGAN} \cite{dcgan} generator as the decoder network and the encoder as the custom 2D neural 
network consisting of five \texttt{ResNet} blocks.
The image latent is the 1024 dimensional vector.
The architectural choice of the augmentation is described in the section about the classification pipeline. We are using the basic method of augmentation by 
inserting an additional memory layer in the decoder network.\footnote{More details on 
the architecture will be given in the released code.}

We observe that the location of the memory layer is important on how the memory is utilized on the train/validation sets and the final reconstruction results.

\chapter{Experiments}

\section{Experiments on random labels}

The heart of our methodology to test the memory layers and re-initialization technique is a well-known variant of randomized test in non-parametric statistics \cite{edgington2007stat, zhang2016und}.
Drawing motivation from \cite{zhang2016und} we have conducted several experiments to test 
the ability of our memory layer to fit the randomly labelled set of data points. 
For this reason, we have prepared a simple data set with \(N\) sample points. We are regulating the number of samples to much the memory size \(M\). This is because our goal was having the one-to-one 
correspondence between the input data and the memory values, i.e. ideally overfitted memory layer. Sample points are the vectors 
uniformly generated in \(\mathbb{R}^{8}\) space, i.e. points in 8
dimensional cube. There are a total of \(m\) classes that are uniformly chosen for each data point. We have experimented with the data
set of 100k data points with 10 classes, consequently setting
\(|\mathcal{M}|\) to 100k also.

Architecturally we have limited ourselves with
the simplest model design with two linear projections before and after the memory layer. It is the basic architecture we could think of with no convolutional neural networks involved.
Moreover, we observe that using convolutional layers allows us to fit the model to the dataset ideally. There is some research on the connection between the multi-head self-attention and convolutional layers 
\cite{cordonnier2020On}, so we have tried to avoid the ambiguity and focused on the fully connected layers as the projections in our network. 

Also to compare our key-value structure with classic dense layers, we have replaced memory access with very wide linear layers and point-wise non-linearity, i.e. \texttt{ReLU}, \texttt{sigmoid}.
As it is described in \cite{cheng2016}, wide layer networks are a powerful memorizers, though 
in \cite{yun2018relu} authors are able to get great memorization for small \texttt{ReLU} networks 
also, with some initialization technique for SGD \cite{robbins2007sgd}. So it was interesting to see how the key-value structure memorization capability can be compared with the
wide dense layers. We have used two fully connected layers with the \texttt{ReLU} in the middle. 
The weight matrix of the layers are set to project the 512-dimensional vector to the 
\(\mathbb{R}^{15k}\) space and after applying the nonlinearity, acting as the discriminant function in the feature space divided by hyperplane decision surfaces, we are projecting the vector back to the
space \(\mathbb{R}^{512}\). This network of two projections and the nonlinearity in the middle is the approximation of our memory layer. This is because the \textit{k}-nn function also acts as the discriminator function, more on this in \cite{signalprocessingbook1995} (Chapter 12).

\begin{figure}[h]
    \centering
    \includegraphics[width=\linewidth]{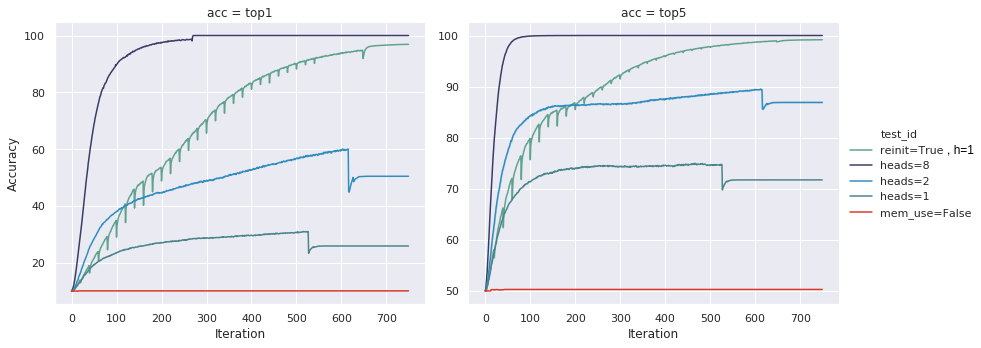}
    \caption{Results for the random data on the various models. The left graph is the plot for top-1 validation 
    accuracy results, top-5 is pictured on the right. We can see that the model with no memory is not able to
    fit the data. Setting 8 heads for multi-head attention memory model, on the other hand, helps the model to easily fit the data. Re-initialization helps to get nearly full convergence with on only 1 head, i.e. multi-head mode disabled. Also, \(h=1\) in the graph is the notation for the single head model.
    }
    \label{fig:random_results}
\end{figure}

We have trained our models with an Adam optimizer \cite{adam}, with an initial learning rate of \(10^{-3}\), and 
\(\beta_1 = 0.9\), \(\beta_2 = 0.98\). 
The models were implemented in \texttt{Pytorch} \cite{pytorch}.
For the memory values we have chosen the \texttt{SparseEmbedding} 
structure which calculates and stores only the sparse gradients in the backward. We have chosen the \texttt{SparseAdam} (implemented in Pytorch) to update the sparse gradients in the memory layer.
Because of the sparse updates in the memory, we have multiplied the learning rate for the sparse parameters by the factor of 10.
For key parameter update, we have used the same optimizer as for the dense 
parameters.
Due to the usage of re-initialization trick and Adam optimizer which stores the values of past gradients and square gradients, these 
values should also be dynamically updated.
The results for the models with memory 
blocks and wide dense layers compared in Figure \ref{fig:random_results}.

In our experiments, we varied the hyper-parameters of the memory model, 
such as memory size, number of heads, k parameter in top-k operator, etc. We provide the results only for \{\(k=10\), \(N=M\)\} hyperparameter set with different values for the number of heads \(h\) and the re-initialization trick enabled/disabled since other combinations 
contain no interest in these experiments.

We observe that setting the number of heads to 8 gives us perfect fit to the data, i.e. full top-1 
validation accuracy. As it is shown in Figure \ref{fig:random_results}, replacing the memory layer 
with wide dense layer doesn't help us with the accuracy. Lowering the number of heads, we see the 
declining accuracy in the validation. We speculate that this is caused by the poor initialization due to which the pair of the same keys could be selected for the two very close query vectors. Using uniform 
initialization to maximize the key coverage at the initial state of the model didn't help us to resolve the 
issue as we have observed that the utilization of keys converged to some small subset.

To overcome the problem, we 
have experimented with \textit{re-initialization trick} that was introduced in the chapter above. As it is seen 
from Figure \ref{fig:random_results}, re-initialization helps us to get nearly ideal validation 
accuracy even with a single head. We are setting \(\epsilon_d=10^{-6}\) to get the results above. We haven't experimented much with 
the special scheduler for the re-initialization trick, but early experiments 
showed that the frequency with which the re-initialization procedure is called and the number of added 
keys for each call can have the significant influence on the final accuracy we get. More experiments are required 
in this direction.

\begin{figure}[h]
    \centering
    \includegraphics[width=\linewidth]{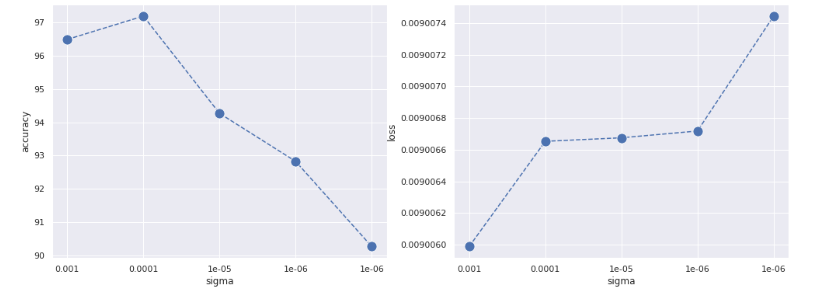}
    \caption{Relation between \(\sigma_d\) hyper parameter of the Gaussian noise square variance parameter in re-initialization procedure and the values
    of train accuracy and loss for the random data. Setting higher \(\sigma_d\) helps significantly, giving lower loss.}
    \label{fig:sigma_results}
\end{figure}

We have conducted some additional experiments to see how the variance \(\sigma_d\) parameter of the 
additive noise added to the re-initialized keys and the memory values affect the final accuracy. We are giving the results in 
in Figure \ref{fig:sigma_results}

\subsection{Results on CIFAR-10}
We have implemented several architectural ideas to test the performance of 
memory augmented models on \texttt{CIFAR-10} \cite{Krizhevsky09learningmultiple}.
The first idea was to augment the bottleneck blocks \cite{he2016deep} with the memory layer and replace the single bottleneck in the network 
with the modified block.
We have also experimented with replacing multiple bottlenecks 
blocks but didn't find anything reasonable to stick with it in the experiments because of 
the overall increased inference time we have observed. 

The logic behind the bottleneck augmentation is given in the chapter above.
Here we describe the architectural choices we've made to incorporate the 
augmented bottleneck in the most optimal way possible taking into consideration the inference time and the final validation accuracy.
The real hurdle during the experiments was the speed issues of the inference. It didn't allow us to set up experiments with more broader set of models because of
the time limitations and the general difficulty of running large grids rapidly for slower models. We were able to partly mitigate the issues by using a lower spatial size of the query input.
Taking all this into the consideration, we have chosen the last 
layer to be augmented with the memory layer as it gave us the smallest spatial size possible in the \texttt{ResNet}-type network.
We have abandoned the experiments with larger spatial sizes in classification experiments for \texttt{CIFAR-10} since the balance between the performance and the accuracy wasn't reasonable. But we still have conducted experiments with larger spatial sizes with the 
autoregressive models, the results are available in the sections below. 

\begin{figure}[h]
    \centering
    \includegraphics[width=\linewidth]{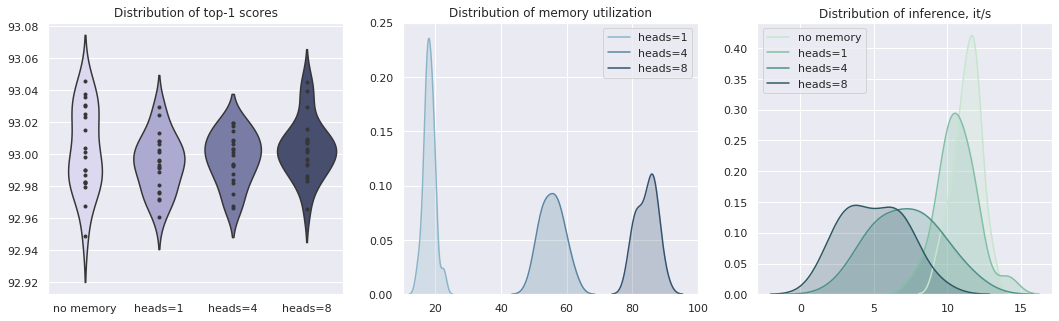}
    \caption{Comparison of the top-1 scores, memory utilization and the inference speed 
    for the model with no memory augmentation and three memory augmented models with the number of heads in the set of \{1, 4, 8\}. 
    There are 20 runs for the each experiment with the different initial seeds. As it can be observed, there is no evident increase in the accuracy while the performance of 
    the models with the heads is much worse compared to the baseline models. Inference values are
    calculated on \texttt{GTX-1080Ti} cards with \textbf{fp32} mode, the results are approximate.}
    \label{fig:cifa10-bottleneck}
\end{figure}

We have chosen the \texttt{ResNet-50} \cite{he2016deep} to be the baseline network 
for the experimental models. The baseline consists of two projections and 16 \texttt{Bottleneck} blocks in the 
middle. We have added the memory layer in the 14th \texttt{Bottleneck} block and have illustrate the 
results in Figure \ref{fig:cifa10-bottleneck}. The training loop design described in \cite{he2016deep} have been implemented with
the SGD \cite{robbins2007ASA} optimizer, learning rate of \(10^{-1}\) weight decay of \(0.0001\) and 
momentum equal to \(0.9\). 
Since the model contained the sparse parameters, we weren't able to use the standard implementation of the SGD optimizer in PyTorch \cite{pytorch}.
For that, we have implemented the \texttt{SparseSGD} optimizer with 
disabled weight decay. As for the momentum, to our knowledge, there is no
mathematical ground of using it to accelerate sparse gradients, but we have
still set it to 0.9 in all of our experiments. More information on the sparse SGD can be found here \cite{ding2019global}.

We have adopted the weight initialization as in \cite{he2015delving} and the batch normalization (BN) 
\cite{batchnorm}. The augmentation is the same as in \cite{he2016deep} with 4-pixel padding on each 
side of the image and 32\(\times\) 32 crop randomly sampled from the padded image and random horizontal 
flip. All of the experiments have been conducted with the memory size of 100k, 
top-\textit{k} operator parameter k of 30 and no \texttt{dropout} on 
the retrieved indices applied. Setting the memory size to 100k we have 
two sets of product keys of size 100, \(|\mathcal{K}_1| = |\mathcal{K}_2| = 100\).

We have calculated the distributions of top-1 scores for 20 runs with different seed values. As it is seen in the Figure \ref{fig:cifa10-bottleneck} we weren't able to gain any 
improvements in the accuracy scores with the memory augmentation, while the performance of the memory model, i.e. iterations per second in the train, decreases significantly and continues to decrease with the higher 
number of heads. We have also calculated the distributions of the memory utilization and observed that for larger heads we see the increase in 
the overall utilization. These findings mirror the results in 
\cite{large_memory}.

\textbf{Evaluation metrics for memory layer}. As the simple evaluation metric of how
well the memory is being utilized during the training phase, we have calculated the memory
usage score which represents the fraction of accessed values \(\#\{\mathcal{C}_i \neq 0\}\),
where \(\mathcal{C}_i \in \mathcal{C}\) is the number of the times the key \(\mathcal{K}_i\)
is accessed summed for the whole validation set. Authors in \cite{large_memory} also use Kullback-Leibler (KL) \cite{Sason_2016} divergence between the distribution of summed values of the \textit{softmax} results for the whole validation dataset with a uniform distribution. We have implemented the KL divergence metric in our experiments and found it giving more accurate numbers with the small changes of the real
memory utilization. But in the given results here we have constrained our experiments to the first 
evaluation metric because of its simplicity and the numerical interpretability.

So as we can see in Figure \ref{fig:cifa10-bottleneck} the utilization of the memory 
is increasing with a larger number of heads. These findings were consistent during all the experiments with the classification networks.

As the results failed on the \texttt{BottleNeck} blocks, we have changed the focus to other architectures. Since we had the problem with the performance due to the large spatial size, 
we have decided to limit ourselves with the image of spatial size \(1\times1\) as the input query for the memory layer. 
Therefore in the next experiments, we have leverage the architectural design of \textit{Squeeze-and-Excitation} \cite{squeeze-e} 
with some changes that were described in the chapter above.

For the experiments with the modified SE blocks, we have 
chosen the \texttt{Resnet-20} as the baseline network. We have kept 
the training pipeline the same but modified the scheduler replacing it with the
\texttt{ReduceLRonPlateau}\footnote{For eg. Pytorch implementation of ReduceLRonPlateau} 
with the reduction factor of \(0.1\). All the experiments with the memory layer enabled have been 
run with a memory size of 100k, top-\textit{k} k parameter of 30 and no dropout on the 
selected indices.
We have listed the most interesting results in the
Table \ref{table:se-blocks-results}

\begin{table}[H]
\begin{center} {\footnotesize
\begin{tabular}{lcccc}
\hline
 & \multicolumn{1}{c}{accuracy, top1} & \multicolumn{1}{c}{inference, ms} & \multicolumn{1}{c}{FLOPs} & \multicolumn{1}{c}{utilization, \%}\\
\hline
\raisebox{0ex}{ResNet-20} & 92.73(92.46$\pm$0.15) & ~6.7ms & $\sim$40.92M & -\\[0ex]
\raisebox{0ex}{SE-ResNet-20} & 93.31(93.16$\pm$0.13)  & ~7.4ms & $\sim$41.49M & -\\[0ex]
\raisebox{0ex}{ResNet-110} & 93.63(93.41$\pm$0.18)  & ~25ms & $\sim$254.98M  & -\\[0ex]
\hline
\raisebox{0ex}{ResNet-20+WL, \(d_w=15k\)} & 92.23 & ~7.2ms & $\sim$43M  & -\\[0ex]
\raisebox{0ex}{ResNet-20+Memory, scalar, h=8} & 92.42 & ~19.5ms & $\sim$41.05M  & 1-2\%\\[0ex]
\raisebox{0ex}{ResNet-20+Memory, cosine, \(\alpha=0.9\), h=8} & 92.45 & ~19.5ms & $\sim$41.05M  & 1-2\%\\[0ex]
\raisebox{0ex}{ResNet-20+Memory/RI, scalar, h=8} & 93.16 & ~19.5ms& $\sim$41.05M & ~60-75\%\\[0ex]
\raisebox{0ex}{ResNet-20+Memory/RI, cosine, \(\alpha=0.9\), h=4} & 93.02 & ~12ms & $\sim$41.05M  & ~40-50\%\\[0ex]
\raisebox{0ex}{ResNet-20+Memory/RI, cosine, \(\alpha=0.9\), h=8} & \textbf{93.51(93.34$\pm$0.14)} & ~19.5ms & $\sim$41.05M  & ~60-75\%\\[0ex]
\hline
\end{tabular} }
\end{center}
\caption{Results for the modified SE blocks. WL is the table notation for the wide linear layer that replaces the
memory layer, \(d_w\) defines respectably the row and column of the projections 
matrices in MLP (with the row vector in the linear operator). Overall we see better results on \texttt{Resnet-20} with the memory layer and with re-initialization trick we have superior 
memory utilization rate. Cosine similarity helps us to nearly reach the accuracy 
values of \texttt{ResNet-100}. Inference values are calculated on \texttt{GTX-1080Ti} 
cards with \textbf{fp32} mode, the results are approximate.}
\label{table:se-blocks-results}
\end{table}

As we can see the re-initialization trick helps us with the utilization of the memory which 
in turn gives us better top-1 accuracy overall. We have also compared the memory block with 
the very wide MLP that consists of two large projections matrices and the pointwise 
nonlinearity in the middle. We are setting the row/column of two matrices to \( d_k =
15\text{k}\), meaning that we have two linear operators \(W_1 \in \mathbb{R}^{d_k\times d_{in}}\) and \(W_2 \in \mathbb{R}^{d_{in}\times d_{k}}\) that map the input vector \(v \in \mathbb{R}^{d_{in}}\) to 
the \(d_k\) dimensional vector, applies \texttt{ReLU} pointwise and project back to the vector \(v^
{\prime} \in \mathbb{R}^{d_{in}}\).
We can see in Table \ref{table:se-blocks-results} that 
adding the large MLP doesn't affect the performance at the level compared to the 
memory layers. It is because the GPUs can easily parallelize the matrix multiplication while stumbling with the operations that require random access to the main memory\cite{kashkovsky2017aspects} . We see this as 
the fundamental problem of the approach with the sparse memories.
 
We haven't conducted experiments with augmenting the \texttt{ResNet-110} network with a memory layer because the goal of these experiments was to understand how the memory layer can help us with the 
very small networks to bit results of large ones. And since the inference speed of the 
small models was inferior compared to \texttt{ResNet} and \texttt{SE-ResNet} blocks we 
have changed our focus to different applications. But more experiments should be conducted to determine whenever \texttt{ResNet-100+M} results compares to the results of \texttt{SE-ResNet-100} both in the 
final prediction scores and the performance.

\begin{figure}[h]
    \centering
    \includegraphics[width=\linewidth]{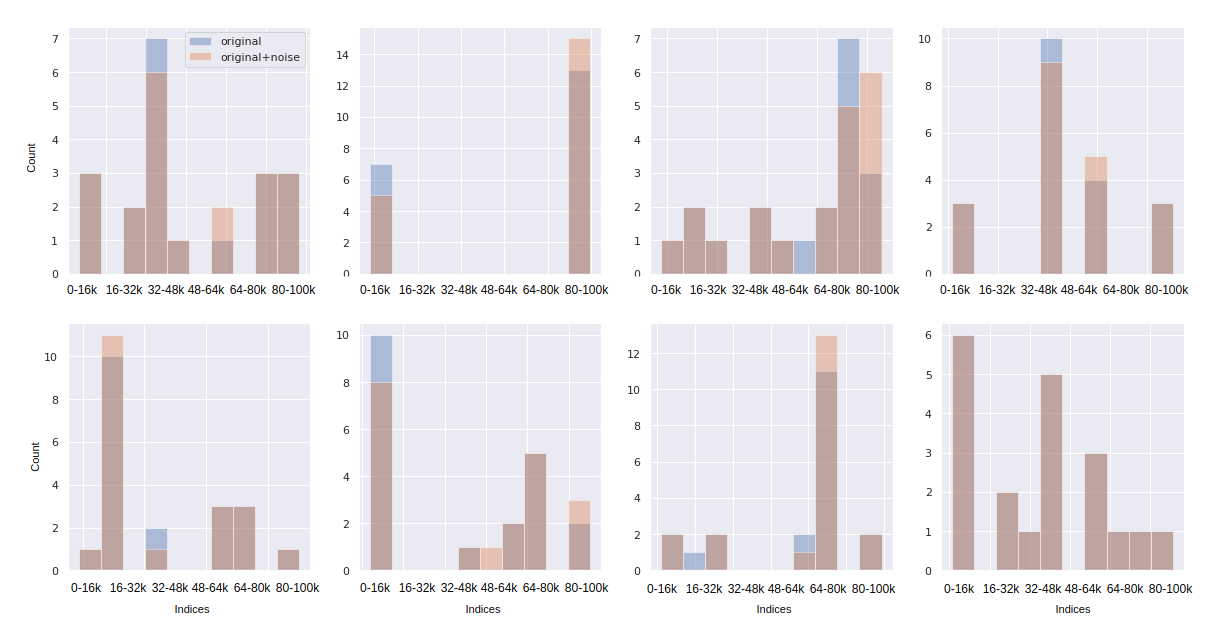}
    \caption{We picture the key access distribution for each head. As we see there is 
    a little perturbation in the distribution when we apply the random region crop in the input image.}
    \label{fig:crop-cifar}
\end{figure}

\textbf{Analysis of memory layer}.
To find how good the introduced memory can generalize to the given images and overall get
the better picture on how the properties of a convolutional neural network, e.g shift-invariance, are maintained with the memory augmentation, we have conducted more experiments in which we
have randomly cropped the small region (4\(\times\)4) of a sample image from the validation and compared the 
accessed keys for the cropped and the original image. We see in Figure
(\ref{fig:crop-cifar}) that the small perturbation of the input data has insignificant 
affect on how the keys selected. Therefore we could assume that the generalization properties of the memory networks are maintained that 
could be crucial in other applications, e.g. pose regression on smaller datasets for which we have conducted additional experiments.

\subsection{Experiments on ImageNet}

We have conducted further experiments on the \texttt{ImageNet-2012} dataset
\cite{russakovsky2015imagenet}, assuming that 
the large size of the train set of the \texttt{ImageNet} would be more natural fit for our 
memory layer. The only issue was time limitations we had and the hard task of tuning 
the optimizer for the memory layer. Since it takes 90 epochs for the \texttt{ImageNet} to finish 
training with \texttt{Resnet-50} and on NVIDIA M40 GPU it takes 14 days \cite{you2017imagenet}, 
the experiments with the size of 224\(\times\)224 weren't reasonable. And since we have decided to
increase the spatial size of the query input in the memory, the inference performance of the
models plummeted. Therefore we have decided to resize the sizes of the images in train and validation to 64\(\times\)64 and run the pipeline. We kept the same augmentation pipeline as in \cite{he2016deep}.

First steps were to run the \texttt{ResNet-50} augmented with the memory layer and compare it with the 
baseline results. The augmentation logic we have chosen for the \texttt{ImageNet} experiments were 
simpler. We have inserted the additional layer before the 44th layer of the network, where the 
image has the 7\(\times\)7 spatial size, this meant that the queries consisted of 49 feature vectors that are batched 
together to be fed to the memory layer. The memory size of the experiments was set to 256k, we 
have looked at the top-30 indices during the memory access and the batch size was set to 256. 
As the distance metric, we have chosen the cosine similarity with \(\alpha=1\). We 
haven't used \texttt{dropout} on the retrieved indices. SGD \cite{robbins2007ASA} was chosen as 
the optimizer with the initial learning rate of \(10^{-1}\), weight decay for dense parameters 
of 0.0001 and momentum of 0.9. We haven't set the weight decay for memory parameters because of the
inferior results, more experiments should be conducted to find the reason for this.

The results are given in \ref{table:resnet-50-imagenet}.

\begin{table}[H]
\begin{center} {\footnotesize
\begin{tabular}{lcccc}
\hline
 & \multicolumn{1}{c}{accuracy, top1} & \multicolumn{1}{c}{utilization}&
 \multicolumn{1}{c}{FLOPs}&
 \multicolumn{1}{c}{inference, ms}\\
\hline
\raisebox{0ex}{ResNet-50, 64\(\times\)64} & 61.45 (61.24$\pm$0.12) & - & $\sim$338M & 11ms\\[0ex]
\raisebox{0ex}{ResNet-50+M, 64\(\times\)64, skip=True,heads=8} & 61.61 & \(\sim\)98\% &$\sim$382M & 23ms\\[0ex]
\raisebox{0ex}{ResNet-50+M, 64\(\times\)64, skip=True,heads=8, \(\text{lr}_{key}\)=\(10^2\)} & \textbf{61.82 (61.59$\pm$0.17)} & \(\sim\)96\%&$\sim$382M& 23ms\\[0ex]
\raisebox{0ex}{ResNet-50+M, 64\(\times\)64, skip=True,heads=8, \(\text{lr}_{key}\)=\(10^3\)} & 60.85 & \(\sim\)86\%&$\sim$382M & 23ms\\[0ex]
\raisebox{0ex}{ResNet-50+M, 64\(\times\)64, skip=False,heads=8} & 54.12 & \(\sim\)15\% &$\sim$382M & 23ms\\[0ex]
\hline
\raisebox{0ex}{ResNet-50, 32\(\times\)32, skip=True} & 42.81 & - & $\sim$86.1M & 6ms\\[0ex]
\raisebox{0ex}{ResNet-50+M, 32\(\times\)32, skip=True,heads=8} & 43.49 & \(\sim\)98\% & $\sim$100M & 12ms\\[0ex]
\raisebox{0ex}{ResNet-50+M, 32\(\times\)32, skip=False,heads=8} & 35.68 & \(\sim\)18\%  & $\sim$100M & 12ms\\[0ex]
\hline
\end{tabular} }
\end{center}
\caption{The results on the \texttt{Resnet-50} and \texttt{Imagenet}. We have tested a number of 
the hyper parameters to find the best train strategy for the memory models. For now we dont see the clear picture on optimization issues.}
\label{table:resnet-50-imagenet}
\end{table}

We can see from the table that there is a small increase in the validation accuracy for the models 
augmented with the memory layer but the large drop in the 
performance (inference in the table). This is not 
a reasonable way of incorporating the memories with 
the classification models and that is why we have tried to analyze how the values in 
the memory were used in the inference and how did they change during the training phase. We hoped to find a way of 
increasing the accuracy of memory augmented models by 
tweaking the training pipeline.
For that we have
logged the gradients of the keys, memory values, memory utilization and standard deviation of the keys during the 
training phase. 

\begin{figure}[H]
    \centering
    \includegraphics[width=\linewidth]{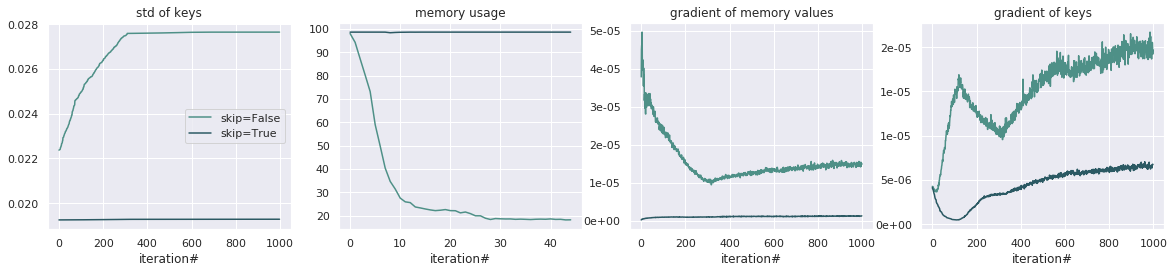}
    \caption{Change in values of the standard deviation of keys, memory usage, gradient of memory values, and key gradients during a training phase.}
    \label{fig:imagenet-graph}
\end{figure}

We have observed that for the activated residual connection on the memory layer, \texttt{skip=True} in Figure 
\ref{fig:imagenet-graph}, gradients were overall higher both for memory and key values. The utilization of the \texttt{skip=True} was way higher reaching almost 100\%, while the \texttt{skip=False} run plummeted to nearly 20\%. What is most interesting is that the standard deviations of the keys in \texttt{skip=True} were not even during all 
training iterations. Our first assumption of the reason for this phenomena was the low learning keys for the key 
parameters. Further experiments are needed to tune the learning rate parameters. As a first step, we have conducted more experiments with super-convergence \cite{smith2019super} to find the top value learning rate for key parameters in a single cycle train. 
We have observed that the super-convergence leaarning rate schedule reaches \(10^6\) before the overall loss starts to 
increase. 

We are not aware of all the underlining issues that do not 
allow us to get the learning rate in a reasonable range. Maybe setting the learning rate value to 
\(10^6\) is logical too, but for now, we don't know that yet.
Also, we require the augmented models to get a way better final accuracy results taking into the consideration the performance issues of the memory blocks and the amount of the additional parameters introduced 
into the network, i.e. \(|\mathcal{M}|\cdot(d_q + d_v)\) where \(d_v\) is the dimension of values in the memory. Because of this we stop our analysis here and acknowledge the need 
for more experiments.

\section{Memory in PoseNet}
We have conducted some experiments on PoseNet \cite{posenet} for 6-DOF camera relocalization.
As it was mentioned in the previous chapter, authors of the paper have used the \texttt{GoogleNet} \cite{googlenet} 
the backbone for feature extraction. We have replaced it in a favour of \texttt{ResNet-34} and run 
several models tuning the hyperparameter set, especially the 
scaling factor \(\beta\). The \texttt{Adam} was the 
optimizer choice in these experiments. We have set the initial learning rate of \(10^{-3}\) 
which decreases every 80 epoch. The weight decay for the dense parameters, i.e. all the parameters
except the key and value vectors in the memory layer, was set to \(2 \times 10^{-4}\). 
In all the results listed, we set weight decay of memory layer to zero,
Setting higher values for the weight decay was the plan of our initial experiments also, as we have hoped to provide some regularization for the values in the 
memory, but even the smallest weight decay failed to give us any 
reasonable results. We acknowledge that the additional work should be done here to 
find the reason behind this issue.
We have set the memory size 
to 1k/10k and compared the results. Overall we have trained the models to the 250th epoch and
have observed the plateau in the train loss.

We have initiated the experiments on the \textit{King's College outdoor relocalization} \cite{posenet}
which is the dataset with the images from Cambridge landmark. There are overall 1220 images in the train set and nearly 
350 in the validation set. The small train set size has discouraged us to apply larger memory sizes
\(|\mathcal{M}|\). Since the validation set is relatively large, we have assumed that the validation accuracy could give us an overview of how good the memory layers generalize to the 
dataset. For the augmentation part, we have resized the images to 256\(\times\)256 and applied a 
random crop of 224\(\times\)224, the same set of transformations have been done in the validation. 
We have set the batch size of the train set to 75 and run the experiments, the results are listed in Table \ref{table:posenet-results}.

\begin{table}[H]
\begin{center} {\footnotesize
\begin{tabular}{lccccc}
\hline
 & \multicolumn{1}{c}{validation} & \multicolumn{1}{c}{train loss} & \multicolumn{1}{c}{inference} & \multicolumn{1}{c}{FLOPs} & \multicolumn{1}{c}{utiliazation}\\
\hline
\(L_{r}(q_1, q_2) = \left\|\mathbf{q}_{1}-\mathbf{q}_{2}\right\|, q^{0}_{i} \in \mathbb{R}^{+}\)\\
\hline
\raisebox{0ex}{PN (PoseNet), $\beta=1$} & 3.02m, 3.17$^{\circ}$ &  37.5 & ~5.6ms & $\sim$3B & -\\[0ex]
\raisebox{0ex}{PN, $\beta=0.1$} & 1.57m, 5.73$^{\circ}$ &  6.57 & ~5.6ms &$\sim$3B & -\\[0ex]
\raisebox{0ex}{PN+LM, $\beta=1$} & 3.24m, 3.43$^{\circ}$ & 34.5  & ~5.8ms &$\sim$(3B+10M) & -\\[0ex]
\raisebox{0ex}{PN+M, heads=1, $\beta=1$, \(|\mathcal{M}|=1k\)} & 3.47m, 3.49$^{\circ}$ & 10.81 
 & ~6ms &$\sim$(3B+16k) & 20-30\%\\[0ex]
\raisebox{0ex}{PN+M, heads=16, $\beta=1$, \(|\mathcal{M}|=1k\), dp=0.3} & 4.43m, 7.23$^{\circ}$ & 14.27 
& ~13ms &$\sim$(3B+16k) & 95-100\%\\[0ex]
\raisebox{0ex}{PN+M, heads=16, $\beta=1$, \(|\mathcal{M}|=1k\)} & \textbf{2.19m},
\textbf{2.99}$^{\circ}$ & \textbf{7.94} & ~13ms &$\sim$(3B+16k) & 95-100\%\\[0ex]
\raisebox{0ex}{PN+M, heads=16, $\beta=0.1$, \(|\mathcal{M}|=1k\)} & \textbf{1.47m},
\textbf{5.02}$^{\circ}$ & \textbf{1.76} & ~13ms &$\sim$(3B+16k) & 95-100\%\\[0ex]
\raisebox{0ex}{PN+M, heads=16, $\beta=0.1$, \(|\mathcal{M}|=10k\)} & \textbf{1.44m},
\textbf{4.98}$^{\circ}$ & \textbf{1.68} & ~15ms &$\sim$(3B+50k) & 50-55\%\\[0ex]
\hline
\(L_{r}(q_1, q_2) = \min \left\{\left\|\mathbf{q}_{1}-\mathbf{q}_{2}\right\|,\left\|\mathbf{q}_{1}+\mathbf{q}_{2}\right\|\right\} \)\\
\hline
\raisebox{0ex}{PN (PoseNet), $\beta=1$} & 2.86m, 3.42$^{\circ}$ &  33.1 & ~5.6ms & $\sim$3B & -\\[0ex]
\raisebox{0ex}{PN, $\beta=0.1$} & 1.59m, 5.36$^{\circ}$ &  6.45 & ~5.6ms &$\sim$3B & -\\[0ex]

\raisebox{0ex}{PN+M, heads=16, $\beta=1$, \(|\mathcal{M}|=1k\)} & \textbf{1.87m},
\textbf{2.53}$^{\circ}$ & \textbf{7.41} & ~13ms &$\sim$(3B+16k) & 80-95\%\\[0ex]

\raisebox{0ex}{PN+M, heads=16, $\beta=0.1$, \(|\mathcal{M}|=1k\)} & \textbf{1.42m},
\textbf{4.38}$^{\circ}$ & \textbf{1.63} & ~13ms &$\sim$(3B+16k) & 80-95\%\\[0ex]
\hline
\end{tabular} }
\end{center}
\caption{Results on the King's College dataset, \(\mathrm{trans}\) and  \(\mathrm{rot}\) are the translation and rotation errors respectively. We see a huge decrease in the loss on the train set for the 
networks with memory augmentation but also the increase in the inference time. Inference values calculated on the \texttt{GTX-1080Ti} with the batch size of 1.}
\label{table:posenet-results}
\end{table}

We have compared the memory networks with the wide MLP layer that is defined as LM in the table.
As in the classification experiments, the MLP layer consists of the two projection matrices and the nonlinearity between. 
First projections matrix maps the input vector \(v_{in} \in \mathbb{R}^{d_{in}}\) to the
\(\mathbb{R}^{2k}\) then applying \texttt{ReLU} on the result we project the vector back to 
\(\mathbb{R}^{d_{in}}\). We are using the residual on the MLP. As it can be seen from the 
table replacing the memory with MLP increases the train and the validation results both for 
rotation and positional losses. For now, we don't understand why the replacement of the memory layer with the MLP can't compete in the final score with the memory block augmentation.

Overall we have seen the huge decrease in the train loss for memory models, while the validation loss, though decreased 
both for rotation and position loss, didn't give us as a steep decrease in the value as we have expected. 
We assume that the more elaborate regularization technique could be applied here.
But for now we have applied the naive dropout regularization on retrieved keys which didn't give us any promising results (dp=0.3 for dropout rate in the Table \ref{table:posenet-results}).

Though getting better numbers overall we are seeing the huge inference time increase for all the models  augmented with the memory.

\subsection{Analysis of the memory layer}

\begin{figure}[H]
    \centering
    \includegraphics[width=\linewidth]{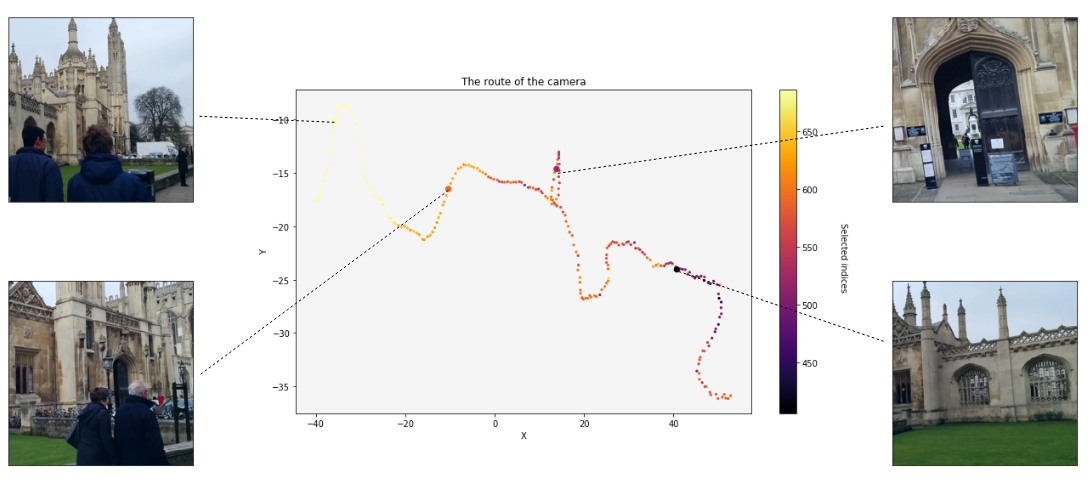}
    \caption{The camera path and the four sample images captures in 
    the given coordinates. We see the correlation between the traveled 
    distance and the indices selected.}
    \label{fig:path}
\end{figure}

To get a better picture of how the memory layer is utilized in regressing the position coordinates, 
we have plotted the distribution of the accessed keys for each image. We have scattered 200 first 
images in the validation set by their x,y coordinates. We have set the colour for each point 
ranging it from 0 to \(|\mathcal{M}|\). To calculate the colour for a particular image we have 
gathered the key indices that were used in the forward operation, averaged and rounded to the nearest integer. The 
results are given in Figure \ref{fig:path}.

We see that there is a correlation between the distance passed by the 
camera and the colours of the point, as they get darker with more distance, i.e. use lower key indices. We could assume 
that the memory can capture the spatial differences between the images and interpret them in the right manner.

\section{Image reconstruction experiments}

We have conducted several experiments to test the performance of the
memory layers in the reconstruction tasks. For that, we have 
constructed a naive encoder-decoder neural network with the 
memory augmentation in the decoder. The overall overview of the architecture is described above. 

We have experiment with the various types of memory placement: right after the latent vector (\texttt{mem\_idx=0}), after the first layer in the 
decoder (\texttt{mem\_idx=1}) and so on. We have used the Adam \cite{adam} optimizer with the initial learning rate of \(10^{-3}\) and 
\(\beta_1 = 0.9\), \(\beta_2 = 0.98\). \texttt{ImageNet} samples were resized to 64 \(\times\) 64 before training the model. 
We have chosen \(L_2\) norm as the objective.
Also, we used the memory size of 100k, k parameter top-\textit{k} procedure of 30 and disabled dropout. The results are given in the Table \ref{table:reconstruction-results}.

\begin{table}[H]
\begin{center} {\footnotesize
\begin{tabular}{lccccc}
\hline
 & \multicolumn{1}{c}{train loss} & \multicolumn{1}{c}{validation loss}& \multicolumn{1}{c}{inference, ms} & \multicolumn{1}{c}{utilization, \%}\\
\hline
\raisebox{0ex}{Baseline} & 0.092 &  0.0076 & ~4.3ms & -\\[0ex]
\hline
\raisebox{0ex}{Baseline+M, heads=1, mem\_idx=0} & 0.0773 &  0.0056 & ~4.3ms & 0-10\\[0ex]
\raisebox{0ex}{Baseline+M, heads=4, mem\_idx=0} & 0.075 &  0.0045 & ~5ms & 10-20\\[0ex]
\raisebox{0ex}{Baseline+M, heads=8, mem\_idx=0} & 0.0734 &  0.0044 & ~6.3ms & 20-40\\[0ex]
\raisebox{0ex}{Baseline+M, heads=1, mem\_idx=1} & 0.0721 &  0.0042 & ~12.1ms & 40-50\\[0ex]
\raisebox{0ex}{Baseline+M, heads=4, mem\_idx=1} & 0.0693 &  0.0037 & ~15.4ms & 50-60\\[0ex]
\raisebox{0ex}{Baseline+M, heads=8, mem\_idx=1} & 0.0675 &  0.0035 & ~17.7ms & 60-65\\[0ex]
\raisebox{0ex}{Baseline+M, heads=1, mem\_idx=2} & 0.0628 &  0.0026 & ~21.4ms & 85-90\\[0ex]
\raisebox{0ex}{Baseline+M, heads=4, mem\_idx=2} & \textbf{0.0544} &  0.0019 & ~26.7ms & 90-100\\[0ex]
\hline
\end{tabular} }
\end{center}
\caption{Reconstruction results after 1600 iterations. The memory utilization numbers are approximate since the amount of experiments conducted was low. The inference speed is calculated on \texttt{GTX-1080Ti} with batch size of 1.}
\label{table:reconstruction-results}
\end{table}

\begin{figure}[H]
    \centering
    \includegraphics[width=\linewidth]{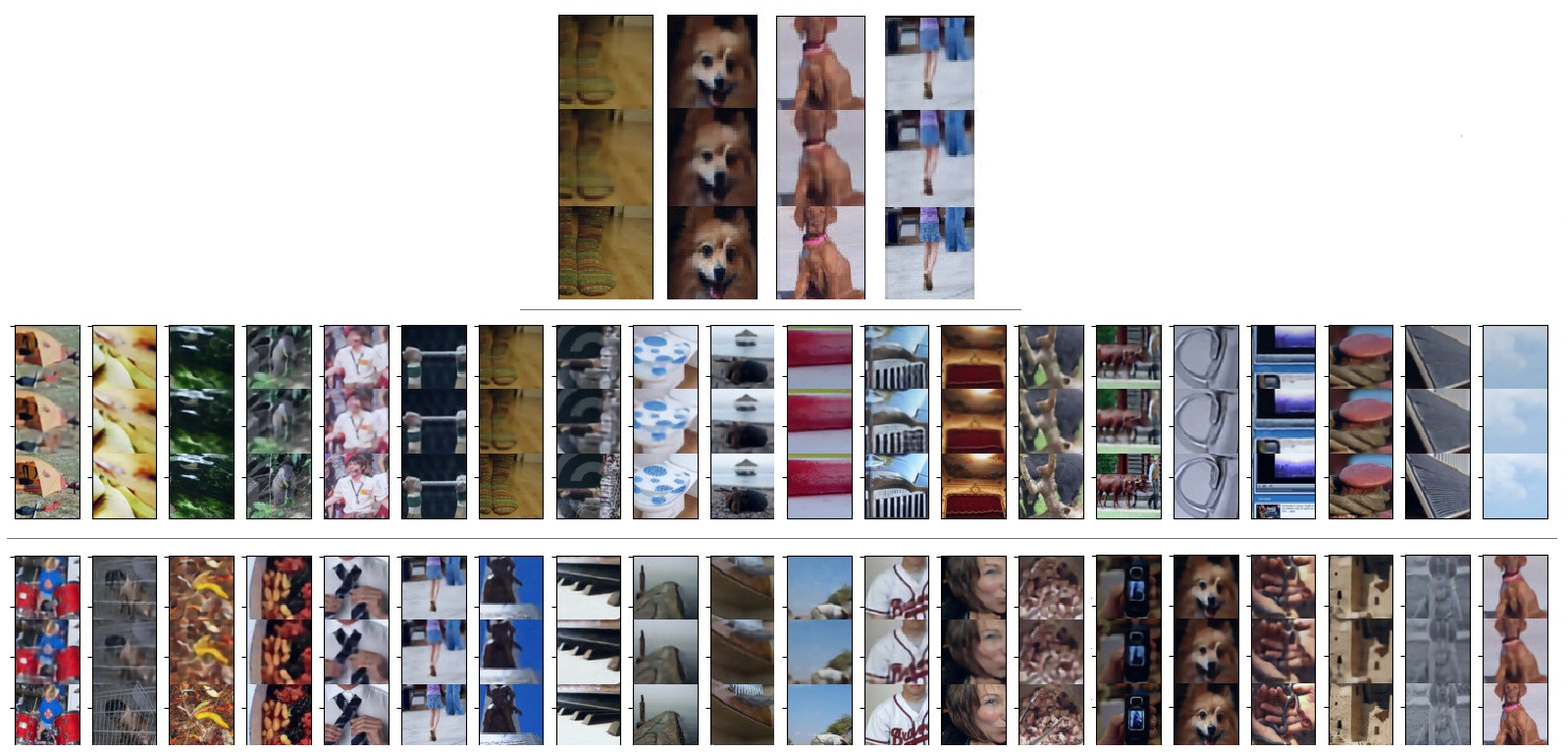}
    \caption{Reconstruction results. Top row images in each section are the reconstruction results with the memory augmented autoencoder  with \texttt{mem\_idx=2} and \texttt{heads=8}, the middle are the output of the baseline autoencoder while the final row are the input images. We see that there are some little details that are 
    captured using the memory layer.}
    \label{fig:reconstruction}
\end{figure}

We see the steady decline in the train and validation losses with 
increasing the number of heads and the index of the layer of the decoder where the memory is being inserted, \texttt{mem\_idx}. 
Utilization numbers increase which again supports the 
experiments we have conducted before.
As the inference time giving us 
the degraded performance with \texttt{mem\_idx=2}. This didn't allow us to conduct more experiments with large spatial shapes of the
input images.
We include some reconstruction examples from the validation set in Figure \ref{fig:reconstruction}.

The overall pipeline and the more details on the final architecture will be given in the released code. For now, it is important to get an understating of the overall reconstruction improvements with the memory augmentation and if it is reasonable to be used with the performance issues in mind.

\section{Other experiments}
We have also applied other experiments with memory usage in distillation \cite{hinton-distill}, implicit shape modelling \cite{chen2019shapemodeling}
and NERF (Representing Scenes as Neural Radiance Fields for View Synthesis) \cite{mildenhall2020nerf}. Overall, for now we can conclude that the large batch sizes of 
these models' training pipelines, i.e. point coordinate samples for implicit modeling and the sampled rays for view synthesis in NERF, is the hurdle which won't allow the memory to be used in the most efficient way, because of the difficulty of the random access parallelization with modern GPUs. Though we see some potential in knowledge distillation from very large models and more work should be conducted in this direction.
\chapter{Conclusion}
This work analyzes the usage of product key-value memory in computer vision applications with a deep dive into the problems like image classification, regression (in the context of the camera relocalization, and the image reconstruction). We have found that for some of the problems the product key memory layer was able to provide significant speed/accuracy improvements with the high utilization of the key-value elements, while others require more careful fine-tuning and an efficient regularization strategy. We also find that the "dying keys" affect the image classification problems. To help us tackle with it, we introduce a simple technique of memory re-initialization which helps us to eliminate "unused" key-value pairs from the memory and cleverly re-initialize new keys that, with high probability, will be used in next iterations of training.

We show that the re-initialization has a huge impact on a toy example of randomly labelled data and observe some gains in performance on the image classification tasks.

In addition to the promising results in the experiments with the camera relocalization, we have also shown that the choice of the set of memory accessed indices in the inference depends on the spatial correlations of the input images. This signals us about the perseverance of the generalization property of the memory layer with no additional regularization required. Still, validation results didn't meet our expectations and at this point, we could only assume that more work is 
required in defining more elaborate regularization strategies.

We hope that the re-initialization training strategy could be applied in large language models with exaggerated number of heads \cite{radford2019language,devlin2018bert,brown2020language}. As we have analyzed, smaller number of heads gave a significant boost in performance.
\begin{singlespace}
\bibliography{main}
\bibliographystyle{plain}
\end{singlespace}

\end{document}